\documentclass{article}

\usepackage{arxiv}
\usepackage{cite}
\usepackage{multirow}
\usepackage{amsmath,amssymb,amsfonts}
\usepackage{algorithmic}
\usepackage{graphicx}
\usepackage{textcomp}
\usepackage{xcolor}
\usepackage{url}
\def\BibTeX{{\rm B\kern-.05em{\sc i\kern-.025em b}\kern-.08em
    T\kern-.1667em\lower.7ex\hbox{E}\kern-.125emX}}

\begin{document}

\title{Probabilistic causal graphs as categorical data synthesizers: do they do better than Gaussian Copulas and Conditional Tabular GANs? \\
}

{}


\author{
Olha Shaposhnyk, Noor Abid, Mouri Zakir, and Svetlana Yanushkevich\\
  Biometric Technologies Laboratory, Schulich School of Engineering\\
  University of Calgary, Canada\\
  \texttt{olha.shaposhnyk1@ucalgary.ca} \\
  }

\maketitle

\begin{abstract}
This study investigates the generation of high-quality synthetic categorical data, such as survey data, using causal graph models. Generating synthetic data aims not only to create a variety of data for training the models but also to preserve privacy while capturing relationships between the data. The research employs Structural Equation Modeling (SEM) followed by Bayesian Networks (BN). We used the categorical data that are based on the survey of accessibility to services for people with disabilities. We created both SEM and BN models to represent causal relationships and to capture joint distributions between variables. In our case studies, such variables include, in particular, demographics, types of disability, types of accessibility barriers and frequencies of encountering those barriers.
	The study compared the SEM-based BN method with alternative approaches, including the probabilistic Gaussian copula technique and generative models like the Conditional Tabular Generative Adversarial Network (CTGAN). The proposed method outperformed others in statistical metrics, including the Chi-square test, Kullback-Leibler divergence, and Total Variation Distance (TVD). In particular, the BN model demonstrated superior performance, achieving the highest TVD, indicating alignment with the original data. The Gaussian Copula ranked second, while CTGAN exhibited moderate performance. These analyses confirmed the ability of the SEM-based BN to produce synthetic data that maintain statistical and relational validity while maintaining confidentiality. This approach is particularly beneficial for research on sensitive data, such as accessibility and disability studies.

\end{abstract}


\section{Introduction}

Collecting and sharing categorical data such as survey results raises privacy concerns, limiting research availability. These problems make it challenging to create accurate AI models. Synthetic data offers a solution by allowing analysis while maintaining individual privacy and addressing data scarcity.

Synthetic data are generated to replicate key characteristics of the original data and are increasingly used to train AI models, support statistical inference, and expand access to data while mitigating privacy and confidentiality risks \cite{[Nikolenko], NIST, Synth}. Privacy refers to the right of individuals to control the disclosure of their personal information, whereas confidentiality involves protecting collected data and ensuring it is used solely for authorized statistical purposes. To address these concerns, organizations like NIST are developing data management standards aimed at preventing breaches of confidentiality \cite{NIST}. Synthetic data provide a practical solution by enabling analysis on data that closely resemble real datasets while safeguarding participant information. Moreover, they help overcome data scarcity, functioning similarly to data augmentation in machine learning by generating diverse training samples. This is particularly beneficial for enhancing causal models, which often require large, representative datasets to ensure consistent and reliable outcomes.

The categorical data we used as a case study are based on the Canadian Survey on Disability \cite{CSD}, which focuses on environmental barriers to accessibility of services to people with disabilities. Those barriers are categorized into four groups: communications, attitude, information and communication technologies (ICT), and physical/transportation. The disability types include hearing, vision, mobility, flexibility, dexterity, learning, pain, mental health, memory, and developmental.
It is important to analyze these barriers and their relationship with demographics, disability type, and the frequency of encountering each barrier to help break down the barriers that may prevent  those with disabilities to have equal access to opportunities,

However, real-world data in the health and social domain is often scarce and sensitive. Attributes such as age, gender and type of disability belong to these categories. Collecting and sharing surveys on a small fraction of the population in a limited geographical area raises privacy concerns, limiting its research availability. These problems make it challenging to create accurate models of data and its causal relationships.


In this study, we investigate selected approaches for generating and analyzing synthetic data that represent statistics on accessibility barriers for people with disabilities and related factors, extracted from surveys. We consider a design of a Bayesian Network (BN) model that intends to represent causal relationships between those barriers and factors \cite{[Zakir]}, and then apply data synthesis using those relationships. The advantage of the BN in data generating is that they capture the joint distributions of the involved variables, thus offering better 'approximations' of the data.

{This paper is structured as follows: Section II presents the most important related works. Section III outlines the data and methodology applied to this study. The experimental study and the results are described in Sections IV and V.  Finally, Section VI concludes the paper.}

\section{Problem Formulation and Contribution}

{The research problem is formulated as follows: what are the best approaches to create synthetic categorical data that preserve real-world datasets' statistical properties and causal relationships between variables while ensuring the confidentiality of the original data? We focus on causality, because it allows for inferences, such as prediction of potential  scenarios given the historical data and current observations, that is, it allows for causal modelling. The generated data should support the complex dependencies and joint distributions observed in the original dataset to ensure their usefulness for analysis and modelling. At the same time, it is important to guarantee the confidentiality of the original participants by minimizing the risk of re-identification and considering social and health data.}

In this work, we consider categorical data on accessibility barriers faced by people with disabilities. To capture the joint distribution of such data, we created causal graph models developed through structural equation modelling (SEM) and expert validation, followed by implementing probabilistic causal graphs and BNs. We utilize these models for data generation. We hypothesize that these models produce data that more closely reflect real-world patterns by effectively capturing dependencies and joint distributions between variables. The generated synthetic datasets are assessed for their quality and usability.

\section{Literature Review}
Synthetic data generation has become important for addressing challenges related to privacy, data scarcity, and class imbalance across various fields. Various techniques have been developed to create realistic synthetic datasets, including methods based on computer graphics \cite{[Yanushkevich-2006]}, 
traditional machine learning algorithms \cite{[Joshi]}, and, more recently, advanced deep learning approaches. These state-of-the-art methods utilize generative machine learning models such as Generative Pretrained Transformers (GPT) \cite{[Luo-2022]}, Generative Adversarial Networks (GANs) \cite{[Goodfellow-2014]}, \cite{park-2018},  Variational Auto-Encoders (VAEs) \cite{[Wan-2017]}, and probabilistic causal models such as BNs \cite{Bao}.

The general principle behind synthetic data generation is to produce numerical data that approximates real-world distributions by learning from actual statistical patterns. 

In agent-based models created and trained on real observations, synthetic data can be randomly generated using the built agent model. A hybrid approach can also be employed, wherein datasets are first constructed based on statistical distributions and subsequently used to generate synthetic data through agent-based modeling.

GANs, originally developed for image synthesis, have found significant acceptance in tabular data mining \cite{Zhao, [Xu], [Sauber]}. For example, GANs have been successfully used to create synthetic clinical data that support data security \cite{[Nicholas]}. Their ability to generate realistic datasets has made them a promising tool in industries such as healthcare, where data privacy is critical. However, while GANs have demonstrated their effectiveness in various applications, their use in the context of accessibility data, especially for people with disabilities, remains relatively unexplored.

Probabilistic causal models, particularly BNs \cite{b1}, are another approach for synthetic data generation \cite{Bao}. BNs represent the relationships between variables through joint probability distributions and conditional dependencies in the form of a directed acyclic graph. The nodes in a BN represent variables, and the edges between them indicate dependencies. The BN-based approach is able to preserve both statistical properties and relationships between variables in the original dataset. This feature makes BNs helpful in generating synthetic data in industries such as healthcare \cite{[Kaur]}, where capturing complex relationships is crucial. While there is a growing use of BNs in the health field, the use of BNs to create synthetic data on accessibility for people with disabilities is an emerging field. BN can reduce the risk of identifying specific individuals, especially in datasets with rare or sensitive attributes, by ensuring that the generated data retains the statistical properties of the original while protecting privacy \cite{Hassan}. BNs have the advantage of explicitly modelling the dependencies between variables. The ability of BNs to capture these relationships makes them particularly valuable for creating complex datasets, such as those related to accessibility, where understanding the relationships between different factors is essential for meaningful analysis. This approach generates high-quality synthetic data that is both statistically accurate and ethically responsible.

While GPT is used mainly to generate text,  GANs and VAE were used to generate video and time series and signals. Most recently, GANs and VAEs were used to generate tabular data, including categorical data, such as medical and health records  \cite{[Apellaniz],[Nicholas]}.
  Categorical data refers to arrays or tables representing statistical data such as demographics or medical data such as vitals recorded over time.



{Existing methods, including GANs and probabilistic models like BNs, have not been comprehensively applied to generating categorical data. They were also not applied to the case study we consider: analysis of factors contributing to accessibility to services for people with disabilities. 


This paper addresses these gaps by leveraging an SEM-based BN to balance the preservation of the original data’s statistical properties and relationships with privacy preservation in the socially sensitive domain of accessibility research.

\section{Methodology}

This section describes the methodology utilized in this study to synthesize data.

 We applied preprocessing techniques to the raw survey data, followed by the implementation of probabilistic causal graphs such as SEM followed by BN. We compare them with alternative methods such as Gaussian Copula and Conditional Tabular GAN (CTGAN)\cite{[Xu]}. The generated synthetic datasets were assessed using statistical metrics, including the Chi-square test, Kullback-Leibler (KL) divergence, and Total Variation Distance (TVD) (Figure \ref{fig:Gen}). 

In this study, BN is both a tool to synthesize the data and to implement inference or prediction of various scenarios or posterior distributions given prior to and during the current observation. We designed two experiments: one to generate data using several approaches, including BN. In the second experiment, we consider a causal graph model representing the accessibility barriers and related factors \cite{[Zakir]} to validate the synthesized data by applying those to conduct inferential reasoning on the model. 

\subsection{Dataset}
 {For this study, we selected a dataset from Statistics Canada's 2022 Canadian Survey on Disability, published in May 2024 \cite{[Data]}. It includes data on 54,000 Canadians with activity limitations due to long-term health conditions.  
The survey covers ten different disability types (such as hearing, vision, mobility, flexibility and so on) and four barrier categories where we focused on interaction barriers with family/services/healthcare, including gender and age.}

{We applied preprocessing techniques to the raw survey data, including data normalization, to convert numerical measures of the frequency of encountering a combination of barriers and disability into probability values. We applied one-hot encoding for nominal attributes, which are categorical attributes without a specific order, such as gender or barrier type.}

\subsection{Data synthesis approaches}

 To synthesize the data on the accessibility barriers, the following methods/libraries were used:

\begin{enumerate}
\item The Synthetic Data Vault (SDV) library {with functions such as}  Gaussian Copula Synthesize, and CTGAN.

\item DataSynthesizer {that has functions} such as Independent Attribute Mode and Correlated Attribute Mode.

\item PyAgrum {that allows to represent and manipulate the BN model, and also perform data synthesis using the models}.
\end{enumerate}


\begin{figure*}
    \centering
    \includegraphics[width=1\linewidth]{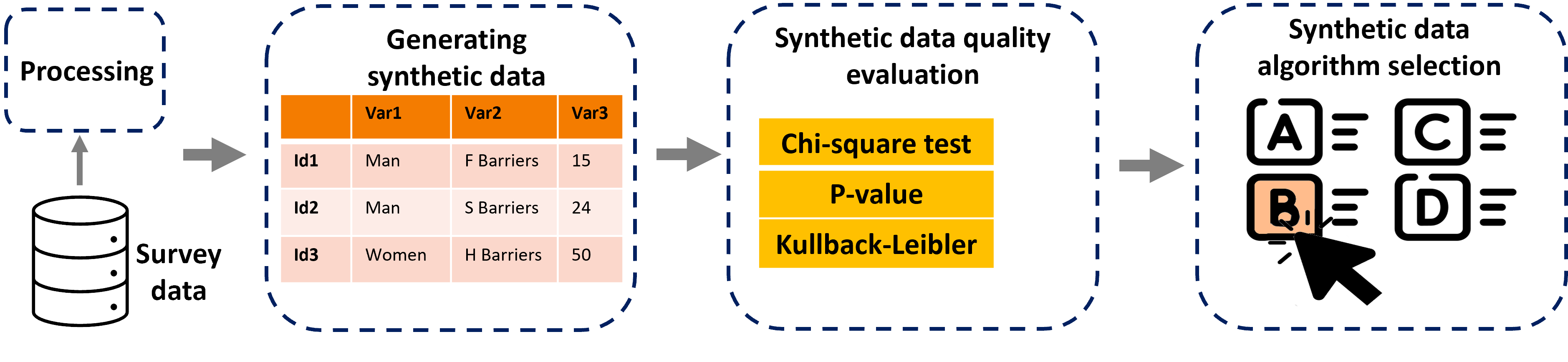}
    \caption{Data synthesis workflow: Preprocessing for format conversion, data generation using various methods (Gaussian Copula, CTGAN, BN, Correlated and Independent Models), evaluation with performance metrics, and selection the optimal algorithm for synthetic data}
    \label{fig:Gen}
\end{figure*}

\subsubsection{Synthetic Data Vault (SDV)}

SDV  \cite{b10} is a Python library designed to facilitate the generation of synthetic data that mirrors the original datasets' format and statistical properties. It supports single-table, multi-table, and time-series data. The SDV provides a variety of models, including both classical statistical methods, such as Gaussian Copula, and advanced deep learning techniques, such as CTGAN. 

Gaussian Copula  \cite{b11} generates synthetic data by modelling the relationships between variables using a statistical model that describes dependencies between variables in a multi-dimensional space. The real data is transformed into a uniform distribution using cumulative distribution functions. The relationships between variables are captured by fitting a Gaussian copula. New data points are generated by sampling from the Gaussian copula model. These samples are then reverse-transformed using the original data's distributions to match the real data's scale and shape.

The CTGAN synthesizer utilizes GAN-based deep learning techniques for model training and synthetic data generation \cite{b10}. CTGAN includes two types of neural networks working together. The Generator creates synthetic data by capturing the patterns and relationships in the real data. The Discriminator evaluates whether a given data point is real or synthetic. Both models are trained together in a loop, improving each other's performance over time.
We tested different settings regarding the number of training epochs (100, 300, 400, 500). More epochs generally improve the Generator's ability to capture complex data patterns but can also increase the risk of overfitting. The batch size was set to 20, which refers to the number of training samples used before updating the model weights. A smaller batch size, such as 20, provides a good balance between computational efficiency and stability. The learning rate was set to 0.0002, which determines the size of the optimizer's steps when updating the model parameters. A small learning rate ensures that the adjustments are gradual for model parameters.

\subsubsection{DataSynthesizer}
DataSynthesizer was proposed in \cite{b12} to generate synthetic datasets from an original private dataset. This includes two types of work: independent attribute mode and correlated attribute mode. Regarding differential privacy, we applied different epsilon values (1, 5, 10).  This parameter ensures the privacy of individuals in a dataset. Lower values provide stronger privacy by injecting more noise, while higher values yield more accurate data with weaker privacy protection. 

The independent attribute mode, used in \cite{b12}, means that each feature in the dataset is generated independently of the others. This means that the synthetic data is generated by learning the marginal distribution of each feature separately without considering how they might be correlated with other features. 

As a part of \cite{b12}, the correlated attribute mode improves upon the independent mode by modelling the correlations between features. Instead of learning the marginal distribution of each feature independently, this model estimates the joint distribution of the features \cite{b12}. The model captures relationships between columns using BN. The BN's depth (number of parents) controls how many dependencies between attributes are considered. We tested 2 and 3 number of parents. The algorithm creates BN automatically based on structure learning and information scores based on provided data.  

\subsubsection{PyAgrum}

To design the BN, we utilized PyAgrum library \cite{b9}. The BN structure explicitly defines the logical and statistical dependencies between variables, using domain knowledge to capture real data dependencies. Once the BN is defined, it can be applied to generate synthetic datasets that preserve the structure and dependencies of the original data. Also, it allows for the inference of the network to compute the probability of specific outcomes given evidence.

Our BN was designed using both the expert knowledge and structural evaluation by the SEM \cite{[Zakir]}. SEM \cite{Igolkina} is a statistical technique that allows for modelling complex relationships among variables, accounting for direct and indirect influences. Its nodes representing demographics (gender, age) are parent nodes to the node corresponding to disability types. Disability type directly impacts interaction barriers like family, services, and healthcare. This helps identify specific challenges and needs based on disability types. The current study focuses only on interaction barriers to simplify our network.

We used an iterative process to create the BN structure. First, we defined it based on expert knowledge. Then, we iteratively employed the SEM framework to ensure the statistical significance and relevance of the structure, confirming that all nodes in the network are statistically important and effectively capture the dependencies within the data.
Once the structure was finalized, we populated the Conditional Probability Tables (CPTs) using the original dataset. These CPTs encode the conditional probabilities of each node given its parent nodes.
 BNs are built on the principles of Bayes' Theorem, joint and conditional probabilities, which enable the network to model complex probabilistic relationships and generate new data consistent with the observed distribution of variables. Bayes' Theorem provides a mechanism for updating the probability of a hypothesis 
\(X\) given new evidence 
\(Y\). The Theorem is given by:

\[
P(X|Y) = \frac{P(Y|X) \cdot P(X)}{P(Y)}
\]

Where:
\begin{itemize}
    \item \(P(X|Y)\) is the posterior probability: the probability of event \(X\) given that \(Y\) has occurred.
    \item \(P(Y|X)\) is the likelihood: the probability of event \(Y\) given that \(X\) is true.
    \item \(P(X)\) is the prior probability: the initial belief about event \(X\) before observing \(Y\).
    \item \(P(Y)\) is the marginal probability: the total probability of event \(Y\), summed over all possible causes.
\end{itemize}

This theorem is crucial for updating beliefs about the state of a variable based on observed evidence. It facilitates the propagation of evidence and supports reasoning under uncertainty, allowing dynamic updates of the network’s beliefs as new data is incorporated.

The conditional probability  \(P(Y|X)\) represents the likelihood of an event \(Y\) occurring given the occurrence of another event \(X\). These relationships are encapsulated in the CPTs, which describe how the presence or absence of one variable influences another within the network.

In a BN, the joint probability of a set of variables is computed as the product of the conditional probabilities for each variable, conditioned on its parent nodes. For a general set of \(n\) variables, the joint probability is computed as:
\[
P(X_1, X_2, ..., X_n) = \prod_{i=1}^{n} P(X_i | \text{Parents}(X_i))
\]

Using this probabilistic framework, we synthesized new data from the BN. The network is able to generate realistic data samples that preserve the statistical relationships observed in the original dataset.

\section{Experiment on Data Synthesis}

This experiment aims to determine whether the SEM-based BN is sufficiently effective in generating synthetic categorical data that best matches the real data from the case study dataset.

The experiment was designed according to the steps presented in  Figure \ref{fig:Gen}. First, preprocessing is performed to convert the raw data into a suitable format for further analysis. This step includes tasks such as cleaning, normalization, one-hot encoding of nominal features, and mapping ordinal data to numerical values. Next, different synthetic data generation techniques were employed, as described in the previous section. Once the data is generated, we evaluate it to determine how closely the synthetic data matches the real data. Finally, the best algorithm for data generation is selected based on the evaluation results.



{To evaluate the quality of synthetic data, we used several key metrics to compare the distributions of the original and synthetic datasets:
\begin{itemize}
    \item Chi-square test and $P$-value: This statistical test measures the sum of the squared differences between the expected and observed counts across categories. Higher values indicate greater differences between the synthesized and original data.
    \item KL divergence \cite{Belov}: The KL-divergence quantifies how much one probability distribution differs from another. A KL value close to zero indicates that the synthetic data is close to the original data distribution, while higher values indicate a greater divergence. Defined as:

    \begin{equation*}
    \centering
    KL = \sum_x P(x) \log\left(\frac{p(x)}{q(x)}\right)
    \end{equation*}


where \textit{p} and \textit{q} are two distribution 

    \item The TVD is a statistical measure used to quantify the difference between two distributions. TVD compares the absolute differences in probabilities between two distributions \cite{Verdu}. TVD value close to one indicates that the synthetic data is close to the original data distribution.
    
    \begin{equation*}
    \centering
    TVD = 1 - \frac{1}{2} \sum_{\omega \in \Omega} |R_{\omega} - S_{\omega}|
    \end{equation*}
    
    where, $\omega $ represents all possible categories in the column ${\Omega}$; $R_{\omega}, S_{\omega}$ are probabilities for each category for $R$ and $S$ datasets.

\end{itemize}
}

\begin{table*}[!ht]
\caption{Comparison of Synthetic Data Generation Methods. Values that show a statistical difference between the distributions for Chi-square statistics ($P$-value  $\ge$ 0.05) are highlighted in bold. Methods were ranked among the best techniques based on evaluation metrics. Methods without statistical significance {were} excluded from ranking}
\centering
\begin{tabular}{l|c|c|c|c|c}
\hline

\textbf{Mode} & {Parameters} & {KL median} & {Chi-Square} &  {TVD}  & {Rank} \\ \hline \hline

Independent attribute mode & {epsilon = 1}& 0.0010 & \textbf{5.1959} & 0.9226 & -\\ 
 & {epsilon = 10}& 0.0003 & \textbf{5.5276} &  0.9746 & -\\
 & {epsilon = 5}& 0.0003 & \textbf{5.5276} &  0.971 & -\\ \hline
 
Bayesian Network & {-} & 3.20E-06 & 2.2564 &  0.9979  & 1\\ \hline

CTGAN & {epochs = 100}& 0.0079 & 4.2157 &  0.9057  &9\\
 & {epochs = 300}& 0.0057 & 1.8822 &  0.9195  & 8\\
 & {epochs = 400}& 0.0057 & 7.9648 &  0.9236  & 6\\ 
 & {epochs = 500}& 0.0071 & 5.0410 &  0.9025 &10\\ \hline
 
Correlated attribute mode: 2 parents & {epsilon = 10}& 0.0008 & \textbf{13.6898}  & 0.9611  & -\\ 
 & {epsilon = 1 }& 0.0052 & 6.7058 & 0.8986 & 11\\
 & {epsilon = 5 }& 0.0013& 3.6500 & 0.9519  & 5\\ \hline

Correlated attribute mode: 3 parents & {epsilon = 10}& 0.0004 & 6.0060 & 0.9684 & 3\\ 
 & {epsilon = 1}& 0.0026& 8.5100 &  0.9208  & 7\\ 
 & {epsilon = 5}& 0.0004 & 4.5879 &  0.9598  & 4\\ \hline

Gaussian Copula & {-}& 0.0002 & 7.1374  & 0.976 & 2\\\hline
\end{tabular}

\label{tab:synthetic_data_comparison}
\end{table*}

\begin{table*}
    \centering
    \caption{Performance Metrics for CTGAN with different numbers of training epochs}
    \begin{tabular}{l|c|c|c}
        {epochs} & {KL median} & {Chi-Square} & {TVD} \\
        \hline \hline
        100 & $0.0102 \pm 0.0029$ & $6.3574 \pm 2.5607$ & $0.9015 \pm 0.0097$ \\ \hline
        300 & $0.0070 \pm 0.0011$ & $3.2457 \pm 1.3874$ & $0.9178 \pm 0.0037$ \\ \hline
        400 & $0.0059 \pm 0.0008$ & $6.1825 \pm 1.8659$ & $0.9222 \pm 0.0058$ \\ \hline
       500 & $0.0066 \pm 0.0011$ & $5.2561 \pm 0.3822$ & $0.9081 \pm 0.0081$ \\ \hline

    \end{tabular}
    \label{tab:synthetic_GAN}
\end{table*}

In this case, the dominating metric is TVD, which is used to make the final decision. All other metrics are auxiliary. We can determine whether our data distributions are statistically significantly similar based on the Chi-square. It helps to exclude data if statistical significance is not achieved for at least one column. The KL can help us decide the best method if the TVD results are similar for several methods.

{The results are summarized in Table \ref{tab:synthetic_data_comparison}, which presents the median chi-square statistics, the median KL divergence values, the TVD, and the rank of each method. So, based on this, we can decide on the similarity of distributions for each variable using different methods of synthetic data generation.}

Our BN model demonstrated the highest TVD of 0.9979, indicating that its distribution closely matches the original data. This suggests that the data generated by the BN is both statistically significant and highly similar to the observed data.  
The Gaussian Copula model ranked second with a TVD of 97.6, showing a strong match to the original data distribution.

For the independent attribute mode, the TVD values ranged between 0.92 and 0.97, with a low $p-value$. While the distributions capture some similarity, they significantly differ from the original data for at least one attribute. 
This result means that the independent generation of attributes introduces discrepancies in the synthetic data, which leads to the model failing statistical validation. Despite adjusting the model parameter epsilon, which marginally increased the TVD, the $p-value$ remained unsatisfactory.

CTGAN provides moderate performance. These methods exhibit TVD values ranging from 0.90 to 0.92. Higher epochs (300, 400) resulted in improved TVD values, indicating a better match to the original data distribution. 
However, the performance slightly declined at epoch 500, suggesting potential overfitting.
The KL Divergence values for these methods range from 0.005 to 0.01, showing that while the methods are effective, the difference between distributions is slightly larger than with methods like BN.

Since the CTGAN model has a stochastic learning process, we have to run the experiment several times to account for variability. Table \ref{tab:synthetic_GAN} presents the performance metrics for CTGAN. Each value is represented as the mean ± standard deviation derived from multiple experiment runs. This approach ensures that the results account for variability and provide a more robust estimate of the synthesizer's performance. Overall, the CTGAN demonstrates stability for our primary metrics (TVD and KL) with a small standard deviation value.

 For the correlated attribute mode, TVD values are between 89 and 96, with the correlated methods typically producing better matches to the original data.
Increasing the number of parents in the correlated mode to three usually improves performance, indicating that adding more parent nodes improves the ability of the model to generate synthetic data that is more similar to the original dataset. 
Changing the mode between the independent to the correlated modes emphasizes the benefit of increased TVD.

{To visualize the resulting distributions, we used a histogram plot. Figure \ref{fig:dens} presents distributions for disability type from the beast synthetic model as BN-generated data and real data.} {This approach allows us to visually assess how closely the synthetic data replicates the real data, capturing key patterns and peaks in the distribution.}

\begin{figure}
    \centering
    \includegraphics[width=0.7\linewidth]{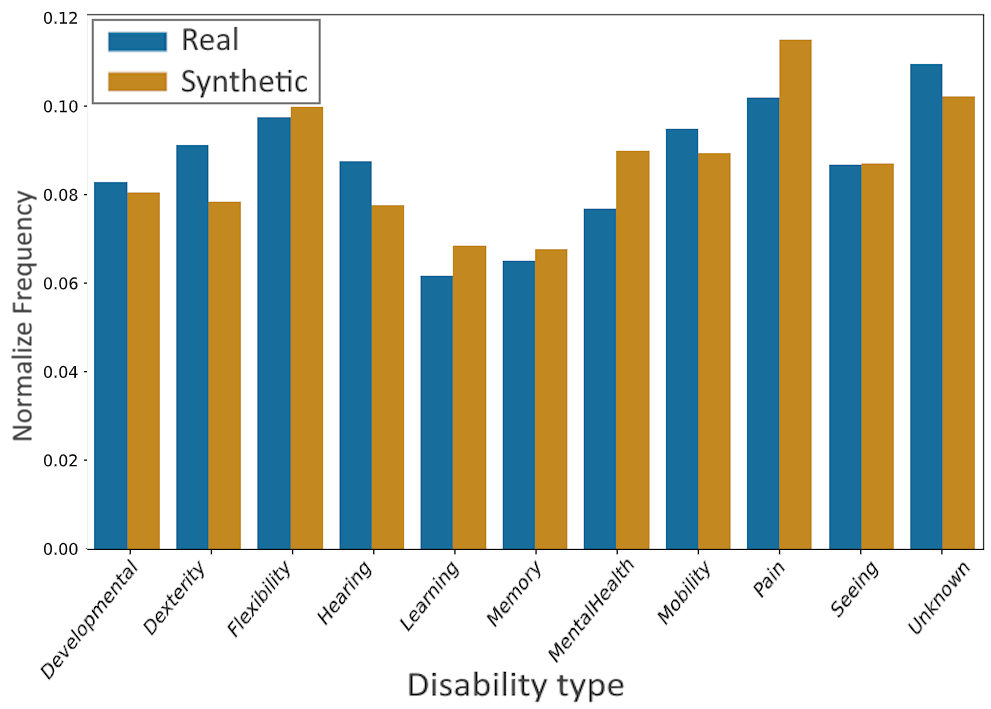}
    \caption{Histogram distribution comparison for disability frequency between real and synthetic data by BN. Real data is highlighted in blue, and synthetic data is highlighted in yellow}
    \label{fig:dens}
\end{figure}

{ 
A good synthetic dataset should not only reflect the marginal distributions of individual variables but also preserve the structural relationships between them. In this context, we focus on metrics like entropy and mutual information to finally evaluate causal graphs created from original data and the best synthetic dataset. Entropy evaluates the variability or uncertainty within each variable independently, while mutual information (MI) captures the dependency or shared information between pairs of variables.}

{Entropy is a metric used to measure the uncertainty or variability within a single variable. When evaluating synthetic data, comparing entropy values between synthetic and real datasets helps determine whether the synthetic version accurately captures the marginal distributions of each variable. In the Table \ref{tab:entropy_comparison}, the entropy values are nearly identical across all nodes. For example, "\textit{Gender}" has an entropy of 0.99237 in the synthetic data compared to 0.99227 in the real data. Similarly, close matches are seen for other variables. These results indicate that the synthetic data preserves the distribution of each feature well. It makes synthetic data a strong candidate for downstream analysis.}

\begin{table}[ht]
\centering
\caption{Comparison of entropy between synthetic and real data across nodes}
    \begin{tabular}{lcc}
    \hline
    \textbf{Node} & \textbf{Entropy Synthetic} & \textbf{Entropy Real} \\
    \hline
        Gender & 0.99237 & 0.99227 \\
        Age & 1.31074 & 1.31236 \\
        Interaction healthcare & 1.95401 & 1.95360 \\
        Disability & 3.43891 & 3.43911 \\
        Interaction services & 1.95936 & 1.95632 \\
        Interaction family & 1.95282 & 1.95211 \\
    \hline
    \end{tabular}
\label{tab:entropy_comparison}
\end{table}

{MI measures the dependency or shared information between pairs of variables. It makes it a useful metric for assessing how well the structure of relationships is retained in synthetic data. By comparing MI values between node pairs in the synthetic and real datasets, we can evaluate whether the synthetic data maintains the same inter-variable associations. In this case, the values shown in Table \ref{tab:mutual_info_comparison} exhibit strong alignment and small differences. These results suggest that the synthetic data preserves the main dependency structure of the original dataset. However, for some nodes relationship such as  ${Gender} \rightarrow {Age}$, MI is very small, which may be due to suboptimal graph structure.}

\begin{table}[ht]
\centering
\caption{Comparison of mutual information between synthetic and real data for node pairs. 
The Source column represents a node where the connection starts.
The Target column represents the node to which the connection is made.}
    \begin{tabular}{cccc}
    \hline
    \textbf{Source} & \textbf{Target} & \textbf{MI Synthetic} & \textbf{MI Real} \\
    \hline
        Gender & Age & 0.001880 & 0.001608 \\
        Age & Disability & 0.019586 & 0.019500 \\
        Disability & Interaction family & 0.070676 & 0.070139 \\
        Disability & Interaction services & 0.062188 & 0.060757 \\
        Disability & Interaction healthcare & 0.057508 & 0.058323 \\
    \hline
    \end{tabular}
\label{tab:mutual_info_comparison}
\end{table}

\section{Application Example of the Synthesized Data to Causal Modelling of Accessibility Barriers}

This part of the analysis explores the use of a causal graph for a part of the decision-making system that represents the causal relationships between accessibility barriers and the factors that cause them. We use the BN model, synthesized privacy data, and the original dataset. Using this data to generate new CPTs for BN nodes, we perform inference on the model using the synthesized data.

The designed BN captures the complex relationships between demographic factors, disability types, and interaction barriers for people with disabilities. The BN require CPT tables to define the probability distribution of each node. 
We created the structure of the BN based on original data, including the relationship between gender, age, type of disability, and type of interaction barrier.

In the BN design using human expert knowledge and evaluated by SEM, nodes representing demographics (gender, age) are parent nodes to the node corresponding to disability types. Disability type directly impacts interaction barriers like family, services, and healthcare. This helps identify specific challenges and needs based on disability types. The current study focuses only on interaction barriers to simplify our network. So, modelling interactions between family, services, and healthcare provides a comprehensive view of how often people may face these barriers. Such a model helps identify which groups are most at risk of experiencing barriers. 

 After the structure was defined, we populated the CPTs using synthetic data to ensure privacy and confidentiality while preserving the statistical properties of the original dataset.

{We show below how the BN can be used to predict or infer potential scenarios given the priors and the current observation. There are two primary ways for the prediction and diagnosis of BN \cite{Fenton}:
\begin{itemize}
    \item Predictive Reasoning: predicting the likelihood of specific outcomes given some observed evidence. 
    \item Diagnostic Reasoning: diagnostics by reasoning backward from observed effects to potential causes.
\end{itemize}
}


\begin{figure*}
    \centering
    \includegraphics[width=0.8\linewidth]{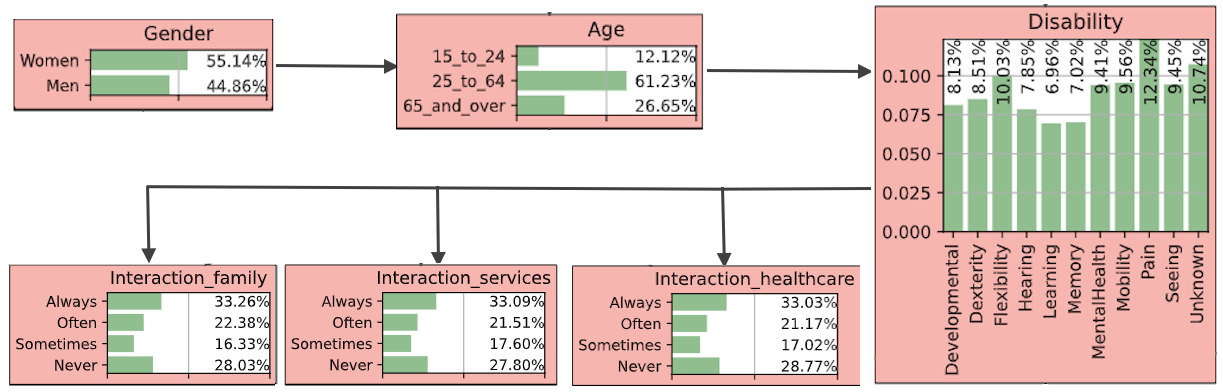}
    \caption{Structure of BN for identifying risks of interaction barriers.}
    \label{fig:BN_synth}
\end{figure*}



Consider two reasoning scenarios.
\paragraph*{Scenario 1}  Consider an individual with a developmental disability (Fig. \ref{fig:bn_part}). Using a predictive case, the BN  estimates the probability of the most likely vulnerable age group for that individual:
  \[P(Age|Disability = Developmental)\]
	This information can help, for example, allocate resources efficiently, such as focused support programs for specific age groups based on predicted vulnerability.
{The results of this inference indicate a change in the age distribution. Before the evidence, the age distribution was heavily weighted toward the adult group  (Fig. \ref{fig:BN_synth}). However, given evidence of developmental disabilities, we observed a shift in the age distribution. Thus, the more vulnerable group is the older people. As illustrated in Figure \ref{fig:bn_part}, the probability for the age group 65 years and older increased significantly to 37\% compared to 26\%.}

\paragraph*{Scenario 2} Using diagnostic reasoning, the BN  assesses the frequency of interaction barriers, such as 
\[P(InteractionFamily|Disability=Developmental)\]
This refers to the likelihood of encountering challenges in family interactions for individuals with developmental disabilities. By identifying underlying problems in interaction, diagnostic considerations can lead to the development of specialized programs such as training for family members.
Let us estimate how often people with \textit{Developmental} disability encounter '\textit{interaction}' barriers. The frequency of family interaction showed a change in trends: after applying the evidence, we observed that the frequency of the 'Often' and 'Always' categories increased. It is very unlikely that they do not face interaction barriers. It suggests that developmental disorders are more common in the adult and oldest age groups and are associated with more frequent family contact. 



\begin{figure*}
    \centering
    \includegraphics[width=0.8\linewidth]{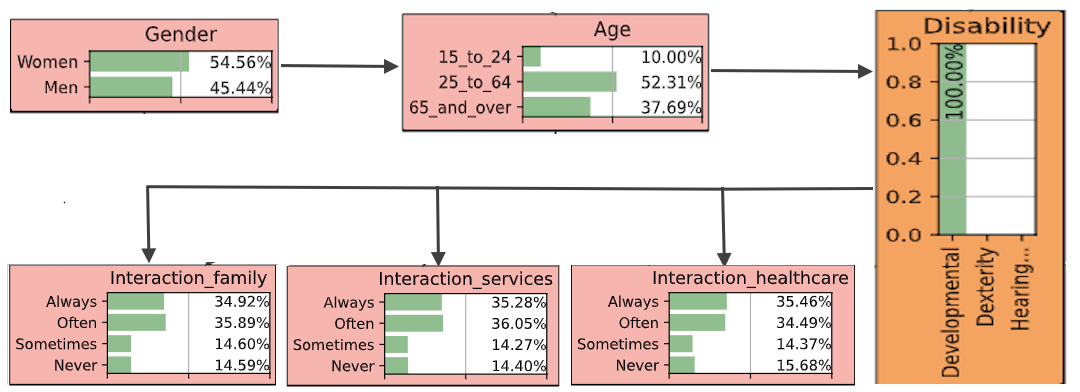}
    \caption{BN, which represents evidence given developmental disability for identifying which groups are most at risk of interaction barriers.}
    \label{fig:bn_part}
\end{figure*}



\section{Conclusion}

This study provides a robust basis for causal inference in decision-making systems, allowing policymakers to make data-driven decisions that support social sustainability. The use of synthetic data in such contexts ensures privacy protection while enabling the development of effective policies and programs aimed at enhancing accessibility and supporting marginalized communities. Our results demonstrate that probabilistic models such as BN incorporating expert validation and structural design generate data reflecting the dependencies between variables.

We evaluated several synthetic data generation methods, focusing on probabilistic causal models,  and compared them against Gaussian copula, as well as generative models such as GAN. We applied preprocessing techniques to the raw survey data, including normalization and encoding, followed by the implementation of SEM-based BN and alternative methods such as Gaussian Copula and CTGAN and others. The generated synthetic datasets were assessed using statistical metrics, including the Chi-square test, KL divergence, and TVD. 
Our evaluation highlights the BN model as the most effective generative approach. It achieved the highest TVD of 0.9979, indicating alignment with the original data. The Gaussian Copula ranked second, while CTGAN exhibited moderate performance. Visual and statistical analyses confirmed the ability of the SEM-based BN to replicate key distributional features of the original dataset. This result confirms the advantages of capturing joint probability distributions and conditional dependencies.


{To validate the synthetic data, we conducted inferential reasoning on the model, the same way BN is used on real data. The synthetic data followed similar distributions yet preserved confidentiality by significantly reducing the uniqueness of records while increasing the overall dataset volume compared to the original dataset.}

We demonstrated the use of the BN in three potential scenarios, using the causal relationships between demographic factors, types of disability, and interaction barriers faced by people with disabilities. 
The BN proved effective for both prognostic and diagnostic purposes, allowing us to understand the most vulnerable demographic groups and the likelihood of encountering specific barriers. 

Including the synthesized data in CPTs ensured confidentiality while preserving the original data set's statistical characteristics. For example, there is a risk of de-anonymization given a small sample size, as individual responses can be traced back to participants. Synthetic data solves this problem, allowing us to analyze while preserving privacy and protecting sensitive information.

Overall, the study demonstrates that causal graph models such as BN do not only capture the joint distributions of variables to allow for causal inference but can also serve as a tool for generating synthetic data that can be used for further model improvement, other models' training and analyses, without disclosing the real data if needed. 


\section*{Acknowledgment}

This project was partially supported by the Natural Sciences and Engineering Research Council of Canada (NSERC).

\section*{Supplementary Data}
The code created for this study can be accessed via the provided GitHub repository link: \url{https://github.com/ExcellentDarkTea/Synthetic-Categorical-Data}

\end{document}